\documentclass{ecai}
\usepackage{graphicx}
\usepackage{latexsym}

\usepackage{times}
\usepackage{soul}
\usepackage{url}
\usepackage[hidelinks]{hyperref}
\usepackage[utf8]{inputenc}
\usepackage{amsmath}
\usepackage{amsthm}
\usepackage{booktabs}
\usepackage{algorithm}
\usepackage{algorithmic}
\usepackage[switch]{lineno}
\usepackage{tabularx}
\usepackage{graphicx}
\usepackage{textcomp}
\usepackage{xcolor}
\usepackage{subfigure} 
\usepackage{amsthm}
\usepackage{url}
\usepackage{float}
\usepackage{multirow}
\usepackage{multicol}
\usepackage{color}
\usepackage{bm}
\usepackage{bbm}

\newtheorem{definition}{Definition}

\usepackage{pifont}

\newcommand{\bp}{\mathbf{p}}
\newcommand{\bx}{\mathbf{x}}

\usepackage{array}
\newcolumntype{L}[1]{>{\raggedright\let\newline\\\arraybackslash\hspace{0pt}}m{#1}}
\newcolumntype{C}[1]{>{\centering\let\newline  \\\arraybackslash\hspace{0pt}}m{#1}}
\newcolumntype{R}[1]{>{\raggedleft\let\newline \\\arraybackslash\hspace{0pt}}m{#1}}
\usepackage{enumitem}
\usepackage{ bbold }

\DeclareMathOperator*{\argmin}{argmin}

\begin{document}
\begin{frontmatter}
	\title{Fair Few-shot Learning with Auxiliary Sets}
	\author[a]{\fnms{Song Wang}\orcid{0000-0003-1273-7694}}
 	\author[a]{\fnms{Jing Ma}\orcid{0000-0003-4237-6607}}
  	\author[b]{\fnms{Lu Cheng}\orcid{0000-0002-2503-2522}}
  	\author[a]{\fnms{Jundong Li}\orcid{0000-0002-1878-817X}}
 \address[a]{University of Virginia}
\address[b]{University of Illinois Chicago}
\begin{abstract}
Recently, there has been a growing interest in developing machine learning (ML) models that can promote fairness, i.e., eliminating biased predictions towards certain populations (e.g., individuals from a specific demographic group). %satisfy specific fairness constraints, such as group fairness. 
Most existing works learn such models based on well-designed fairness constraints in optimization.
Nevertheless, in many practical ML tasks, only very few labeled data samples can be collected, %it can be challenging to obtain sufficient data samples. % for learning a fair model. 
%With insufficient samples, the model performance regarding fairness will inevitably deteriorate due to two reasons: insufficient labeled samples and imbalanced sensitive attributes.
which can lead to inferior fairness performance. This is because existing fairness constraints are designed to restrict the prediction disparity among different sensitive groups, but with few samples, it becomes difficult to accurately measure the disparity, thus rendering ineffective fairness optimization.
%%%%%due to two reasons: insufficient samples for fairness constraints and imbalanced sensitive attributes.
In this paper, we define the fairness-aware learning task with limited training samples as the \emph{fair few-shot learning} problem. To deal with this problem, we devise a novel framework that accumulates fairness-aware knowledge across different meta-training tasks and then generalizes the learned knowledge to meta-test tasks. To compensate for insufficient training samples,
%%%%%%overcome the challenges resulting from insufficient samples,
%%%%%%and imbalanced sensitive attributes, 
we propose an essential strategy to select and leverage an \emph{auxiliary set} for each meta-test task. These auxiliary sets contain several labeled training samples that can enhance the model performance regarding fairness in meta-test tasks, thereby allowing for 
%In this way, the learned useful fairness meta-knowledge can be transferred to meta-test tasks.
the transfer of learned useful fairness-oriented knowledge to meta-test tasks.
Furthermore, we conduct extensive experiments on three real-world datasets to validate the superiority of our framework against the state-of-the-art baselines. 
\end{abstract}
\end{frontmatter}
\section{Introduction}

Machine learning (ML) tools have been increasingly utilized in high-stake tasks such as credit assessments~\cite{moro2014data} and crime predictions~\cite{lichman2013uci}. Despite their success, the data-driven nature of existing machine learning methods makes them easily inherit the biases buried in the training data and thus results in predictions with discrimination against some sensitive  groups~\cite{stevenson2018assessing}. Here, sensitive groups are typically defined by certain sensitive attributes such as race and gender  \cite{sweeney2013discrimination,bolukbasi2016man,buolamwini2018gender,hardt2016equality,zafar2017fairness}.
%often leads to discriminatory (i.e., unfair) predictions~\cite{sweeney2013discrimination,bolukbasi2016man,buolamwini2018gender}. In particular, machine learning models can easily inherit the bias that exists in the historical data used for training, which thus leads to biases in predictions~\cite{hardt2016equality,zafar2017fairness}. 
For example, a criminal risk assessment model can unfavorably assign a higher crime probability for specific racial groups~\cite{stevenson2018assessing}. 
%With the emergence of widely-used machine learning models, 
In fact, such undesirable biases commonly exist in various real-world applications such as toxicity detection \cite{cheng2022bias}, recommendation systems~\cite{lambrecht2019algorithmic}, loan approval predictions~\cite{sarkar2020mitigating}, and recruitment~\cite{dastin2018amazon}.

In response, a surge of research efforts in both academia and industry have been made for developing fair machine learning models~\cite{corbett2017algorithmic,cheng2021socially}. These models have demonstrated their ability to effectively mitigate unwanted bias in various applications~\cite{barrett2019adversarial,zhang2018mitigating}. 
%%%Despite the promising performance of these methods
Many fair ML methods \cite{chuang2021fair,cotter2019training} incorporate fairness constraints 
%designed w.r.t. specific fairness notions such as demographic parity (DP) \cite{grgic2016case}. Generally, these fairness constraint 
to penalize predictions with statistical discrepancies among different sensitive groups.
These methods often rely on sufficient training data from each sensitive group (e.g., collecting data from a specific region with an imbalanced population composition~\cite{zhao2020primal}). However, in many scenarios, only very few data samples can be collected, especially for those from the minority group. This could render existing fair ML methods ineffective or even further amplify discrimination against the minority group. 
%However, they typically tend to underperform 
%%%%when applied to a novel domain with different data distributions, since the fairness constraints are
%when available labeled samples are limited, since the fairness constraints are
%data-dependent, i.e, they are estimated from existing samples~\cite{chuang2021fair,cotter2019training}. 
%%%%While fine-tuning the model based on the novel domain is a viable solution, its effectiveness is limited when the novel labeled samples are scarce.
%%%This is particularly challenging when few labeled samples of the new domain are present, as fine-tuning the model based on the new domain can be ineffective with few labels. 
%%%with few labeled samples. 
%These methods generally assume that (1) there exist sufficient labeled samples for model training and (2) the training data follows a similar distribution to data in the deployment. However, collecting data is often challenging in high-stake domains such as healthcare and recidivism prediction~\cite{christin2017algorithms}, especially the sensitive attributes, due to privacy concerns and legal regulations. 
%For example, a fair criminal risk assessment model trained on data collected from one region may still produce discriminatory results when applied to a new region with only a few labeled samples \cite{lichman2013uci}. 
To enhance the applicability of fair ML in practice~\cite{zhao2020primal}, this work aims to address the crucial and urgent problem of \emph{fair few-shot learning}: promoting fairness in few-shot learning tasks with a limited number of samples.
%developing fair machine learning models that can maintain desired fairness performance with a limited number of samples.

\vspace{0.05in}
One feasible solution to address fair few-shot learning is to incorporate fairness techniques into few-shot learning methods.
%to mitigate data bias under the few-shot scenario. 
Particularly, we first learn from \emph{meta-training tasks} with adequate samples~\cite{snell2017prototypical,finn2017model,vinyals2016matching}, and then leverage the learned knowledge and fine-tune the model on other disjoint \emph{meta-test tasks} with few samples based on fairness constraints. We define such a step of fine-tuning as \emph{fairness adaptation}.
However, there still remain two primary challenges for our problem. %First, the fairness generalization gap between training and testing data can be large, i.e., the data distribution during testing can greatly deviate from that during training. This can lead to largely degraded fairness performance when the model is actually deployed. For example, a fair treatment prediction model trained on data from a healthcare center for the elders may lose the fairness property when applied to a children's hospital with few patient samples. 
First, the insufficiency of samples in meta-test tasks can result in unsatisfactory fairness adaptation performance. Although the model can adapt to meta-test tasks with limited samples via fine-tuning for classification, these samples may not be sufficient to ensure fairness performance. Many fairness constraints are designed to restrict the prediction disparity among different sensitive groups. However, in fair few-shot learning,
%This is because existing fairness constraints aim to restrict the disparity between sensitive groups,  
the lack of samples in each sensitive group inevitably increases the difficulties in measuring the prediction disparity. %leading to suboptimal fairness adaptation. 
Moreover, in meta-test sets, the sensitive attributes of data samples can often be extremely imbalanced  (e.g., a majority of individuals belonging to the same race, while other sensitive groups have very few, or even no samples). In these cases, the conventional fairness constraints are often ineffective, or completely inapplicable. %thus further magnifying the challenge in fairness adaptation.
%%%This is because fairness learning methods are generally proposed for scenarios with sufficient labeled samples and thus inevitably yield inferior performance in few-shot scenarios.
%Second, existing fairness learning methodologies cannot be directly applied to fair few-shot learning due to the potential absence of different sensitive attributes during test.
%%%%%Second, the imbalanced distribution of sensitive attributes can pose challenges in performance fairness adoption. In particular, due to the scarcity of labeled samples, a specific value of sensitive attributes will probably dominate most labeled samples (e.g., a majority of individuals belonging to the same race). For example, the data collected from a specific region may have an overrepresentation of a specific race due to the population composition of that region. In such cases, the effectiveness of fairness adaptation will be reduced.
Second, the generalization gap between meta-training tasks and meta-test tasks hinders the efficacy of fairness adaptation. Similar to other few-shot learning studies, the key point of fair few-shot learning is to leverage the learned knowledge from meta-training tasks to facilitate the model performance on meta-test tasks with few samples. In our problem, it is essential to leverage the learned knowledge for fairness adaptation. 
%Particularly, the fairness adaptation should utilize learned knowledge to compensate for limited samples.
However, models that manage to reduce disparities on meta-training tasks do not necessarily achieve the same performance in fairness on meta-test tasks~\cite{cotter2019training}, due to the fact that fairness constraints are data-dependent and thus lack generalizability~\cite{chuang2021fair}. As a result, it remains challenging to extract and leverage the learned knowledge that is beneficial for fairness adaptation.

%In particular, fair few-shot learning can easily encounter the problem of missing sensitive attributes (e.g., most individuals are of the same gender). For example, due to privacy concerns, a children's hospital may only gather limited samples of the same gender, which can greatly hinder the fair learning process. In consequence, the prevalent fairness constraints that seek a balance between two or more groups will become ineffective in fair adaptation. 
\vspace{0.05in}
To tackle these challenges, we devise a novel framework for fair few-shot learning, named FEAST (\textbf{\underline{F}}air f\textbf{\underline{E}}w-shot learning with \textbf{\underline{A}}uxiliary \textbf{\underline{S}}e\textbf{\underline{T}}s). Specifically, 
we propose to leverage an \emph{auxiliary set} for each meta-test task to promote fair adaptation with few samples while addressing the issues caused by insufficient samples. %and imbalanced sensitive attributes. 
The auxiliary set is comprised of several samples from meta-training data and is specific to each meta-test task.
%%%%%, which serve as additional labeled samples that. 
By incorporating these auxiliary sets via a novel \emph{fairness-aware mutual information loss}, the model can be effectively adapted to a meta-task with few samples while preserving the fairness knowledge learned during training. Furthermore, to effectively leverage the learned knowledge from meta-training tasks for fairness adaptation, %to address the problem of imbalanced sensitive attributes, 
our proposed framework selects the auxiliary sets based on the \textit{fairness adaptation direction}. This ensures that the selected auxiliary sets share similar fairness adaptation directions and thus can provide beneficial learned knowledge.
%%%%%, even when working with few samples that have the same sensitive attributes.
%the fine-tuning process can benefit from the fairness constraints to ensure fair learning. In summary, our contributions are as follows:
We summarize our main contributions as follows:

\begin{itemize}
    \item \textbf{Problem.} We study the crucial problem of fair few-shot learning. We introduce the importance of this problem, analyze the challenges, and point out the limitations of existing studies. To the best of our knowledge, this is the first work that addresses these unique challenges in fair few-shot learning.
    %identify the limitations of existing fair learning methods in scenarios with limited labeled samples and highlight the necessity of conducting fair few-shot learning.
    \item \textbf{Method.} We develop a novel fair few-shot learning framework that (1) can leverage auxiliary sets to aid fairness adaptation with limited samples, and (2) can select auxiliary sets with similar optimization directions to promote fairness adaptation.
    \item \textbf{Experiments.} We conduct extensive experiments on three real-world fairness datasets under the few-shot scenario and demonstrate the superiority of our proposed framework in terms of fairness compared with a couple of state-of-the-art baselines.
\end{itemize}

\section{Problem Statement}
%\subsection{Preliminary}
In this section, we provide a formal definition for the problem of fair few-shot learning that we study in this paper.
Denote $\mathcal{Z}=\mathcal{X}\times\mathcal{Y}$ as the input space, where $\mathcal{X}\subset\mathbb{R}^n$ is the input space with $n$ different features and $\mathcal{Y}=\{1,2,\dotsc,N\}$ is the label space with $N$ discrete classes. We consider inputs $X\in\mathcal{X}$, labels $Y\in\mathcal{Y}$, and sensitive attribute $A\in\{0,1\}$.
In the few-shot setting, the dataset $\mathcal{D}$ is comprised of two different smaller datasets: meta-training data $\mathcal{D}_{tr}$ and meta-test data $\mathcal{D}_{te}$. Moreover, $\mathcal{D}=\mathcal{D}_{tr}\cup\mathcal{D}_{te}$ and $\mathcal{D}_{tr}\cap\mathcal{D}_{te}=\emptyset$, i.e., $|\mathcal{D}_{tr}|+|\mathcal{D}_{te}|=|\mathcal{D}|$. In general, few-shot settings assume that there exist sufficient samples in $\mathcal{D}_{tr}$, while samples in $\mathcal{D}_{te}$ are generally scarce~\cite{finn2017model,sung2018learning}. 
%%%%%In the few-shot setting, the dataset $\mathcal{D}$ consists of many smaller datasets collected from $|\mathcal{D}|$ different subsets, i.e., $\mathcal{D}=\{d_1, d_2,\dotsc, d_{|\mathcal{D}|}\}$. The distribution of subset $d_i$ is denoted as $P_{d_i}(X, Y)$. In practice, some subsets may contain abundant samples whilst others may only contain a significantly smaller number of samples. For example, in the criminal risk assessment dataset~\cite{lichman2013uci}, the samples are collected from different states in the U.S. %%%%%That being said, the numbers of samples in different datasets may vary massively.
%%%%%Hence, 
%%%%%We further divide these subsets into two different categories: the meta-training subsets $\mathcal{D}^{b}=\{d_1^b,d_2^b, \dotsc, d^b_{|\mathcal{D}^b|}\}$ and the meta-test subsets $\mathcal{D}^{n}=\{d_1^n,d_2^n, \dotsc, d^n_{|\mathcal{D}^n|}\}$. Note that $\mathcal{D}=\mathcal{D}^b\cup\mathcal{D}^n$ and $\mathcal{D}^b\cap\mathcal{D}^n=\emptyset$, i.e., $|\mathcal{D}^b|+|\mathcal{D}^n|=|\mathcal{D}|$. General few-shot settings assume that there exist sufficient samples in $\mathcal{D}^b$, while samples in $\mathcal{D}^n$ are generally scarce~\cite{finn2017model,sung2018learning}. Subsequently, our goal is to develop a machine learning model that can accurately and fairly predict labels for samples in $\mathcal{D}^n$ with limited samples after training on $\mathcal{D}^b$. 

%\subsection{Problem Statement}
The proposed framework is built upon the prevalent paradigm of episodic meta-learning~\cite{sung2018learning,snell2017prototypical}, which has demonstrated superior performance in the field of few-shot learning~\cite{finn2017model,vinyals2016matching}. The process of episodic meta-learning consists of meta-training on $\mathcal{D}_{tr}$ and meta-test on $\mathcal{D}_{te}$. During meta-training, the model is trained on a series of \emph{meta-training tasks} $\{\mathcal{T}_1,\mathcal{T}_2, \dotsc, \mathcal{T}_T\}$, where each meta-training task contains support set $\mathcal{S}$ as the reference and a query set $\mathcal{Q}$ to be classified. $T$ is the number of meta-training tasks. More specifically, $\mathcal{S}=\{(x_1,y_1),(x_2,y_2),\dotsc,(x_{N\times K},y_{N\times K})\}$ contains $N$ classes and $K$ samples for each of these $N$ classes (i.e., the $N$-way $K$-shot setting). Meanwhile, the query set $\mathcal{Q}=\{(x^q_1,y^q_1),(x^q_2,y^q_2),\dotsc,(x^q_{|\mathcal{Q}|},y^q_{|\mathcal{Q}|})\}$ consists of $|\mathcal{Q}|$ different samples to be classified from these $N$ classes. 
 Subsequently, our goal is to develop a machine learning model that can accurately and fairly predict labels for samples in $\mathcal{D}_{te}$ with limited labeled samples after training on $\mathcal{D}_{tr}$. Formally, the studied problem of fair few-shot learning can be formulated as follows.
	\begin{definition}
	\textbf{Fair few-shot learning:} Given meta-training data $\mathcal{D}_{tr}$ and a meta-test task $\mathcal{T}=\{\mathcal{S}, \mathcal{Q}\}$ sampled from meta-test data $\mathcal{D}_{te}$, our goal is to develop a fair learning model such that after meta-training on samples in $\mathcal{D}_{tr}$, the model can accurately and fairly predict labels for samples in the query set $\mathcal{Q}$ when the only available reference is the limited samples in the support set $\mathcal{S}$.
	\end{definition}

Note that the support sets and the query sets are sampled from meta-training data $\mathcal{D}_{tr}$. That is, for any sample $(x_i,y_i)$ in a meta-training task, $(x_i,y_i)\sim P_{tr}(X,Y)$, where $P_{tr}(X,Y)$ is the meta-training task distribution from meta-training data $\mathcal{D}_{tr}$. We then evaluate the model on a series of meta-test tasks, which share the same structure as meta-training tasks, except that the samples are now from meta-test data $\mathcal{D}_{te}$. In other words, for any sample $(x_i,y_i)$ during meta-test, we have $(x_i,y_i)\sim P_{te}(X,Y)$, where $P_{te}(X,Y)$ is the meta-test task distribution from meta-test data $\mathcal{D}_{te}$. Under the meta-learning framework~\cite{finn2017model,zhou2019meta,huang2020graph}, the model needs to be first fine-tuned for several steps (i.e., fairness adaptation) using the support set, and then performs fair classification for samples in the query set.

%%%%%Formally, denoting the meta-task distribution in a specific subset $d$ as $\mathcal{T}\sim P_{d}(\mathcal{T})$, the overall meta-training task distribution can be represented as $P_{\text{train}}=\sum_{d\in \mathcal{D}^b}w_d^{b}P_d(\mathcal{T})$. Similarly, the meta-test task distribution is $P_{\text{test}}=\sum_{d\in \mathcal{D}^n}w_d^{n}P_d(\mathcal{T})$. Here $w_d^{b}$ and $w_n^{n}$ are the weights for each meta-training and meta-test subset, respectively.

\section{Proposed Framework}
\begin{figure*}[htbp]
	    \centering
	    \includegraphics[width=0.9\textwidth]{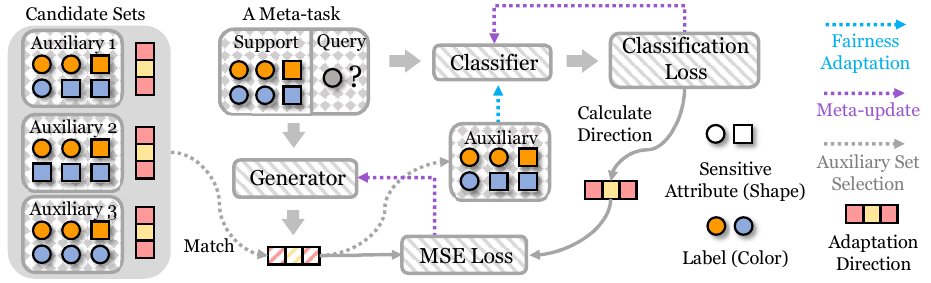}

\caption{The overall framework of FEAST. Here different shapes denote different sensitive attributes, and colors represent sample classes. Given a meta-task, the generator will output the estimated fairness adaptation direction, which is used to select an auxiliary set with the most similar direction from the candidate set. Then we conduct fairness adaptation with the auxiliary set on the current meta-task and perform predictions. The resulting fairness adaptation will be used to update the generator. Note that during training, the meta-task will be incorporated into the candidate auxiliary sets after the optimization of one episode.
}
\label{fig:illustration}

	\end{figure*}

We formulate the problem of \emph{fair few-shot learning} in the $N$-way $K$-shot meta-learning framework. The meta-training process typically involves a series of randomly sampled meta-training tasks, each of which contains $K$ samples for each of the $N$ classes as the support set, along with several query samples to be classified. Under the few-shot scenario, it is challenging to conduct fairness adaptation on the support set due to the insufficiency of samples and the generalization gap between meta-training tasks and meta-test tasks. Therefore, as illustrated in Fig.~\ref{fig:illustration}, we propose the use of auxiliary sets that can enhance fairness adaptation for each meta-test task. In this section, we first introduce the process of conducting fairness adaptation with auxiliary sets and then discuss the strategy to select auxiliary sets. 
%%%%%the model will be severely deteriorated by two problems: insufficient labeled samples and imbalanced sensitive attributes. To overcome these difficulties, we propose to leverage auxiliary sets to enhance fairness adaptation on any meta-tasks.
%%%The overview of the proposed framework is described in Fig.

\subsection{Fairness Adaptation with Auxiliary Sets}
To alleviate the issue of ineffective fairness adaptation to meta-test tasks caused by insufficient samples, we propose to leverage the samples in meta-training tasks for fairness adaptation. 
%Specifically, under the few-shot scenario, since each class in a meta-test task only consists of $K$ support samples, the specific sensitive attributes can be insufficient or oven disappearing. 
%For example, under the 2-way 5-shot setting, each meta-test task only contains 10 support samples in total, which means if there exist two different sensitive attributes, one of the attributes can be potentially insufficient or even lacking among these 10 support samples. Therefore, it is crucial to leverage existing samples in meta-training tasks to compensate for the lack of sensitive attributes in meta-test tasks. 
Specifically, considering a target meta-test task $\mathcal{T}=(\mathcal{S},\mathcal{Q})$, our goal is to utilize an auxiliary set $\mathcal{A}$ obtained from meta-training data that can compensate for inadequate samples in $\mathcal{S}$.
%alleviate the problem caused by the lack of labeled samples and sensitive attributes in $\mathcal{T}$. 
However, due to the distribution difference between meta-training tasks and meta-test tasks, it remains non-trivial to leverage the auxiliary set $\mathcal{A}$, which follows a different distribution from $\mathcal{S}$. Since the data distribution in $\mathcal{A}$ differs from that in $\mathcal{S}$, directly conducting fairness adaptation on $\mathcal{A}$ can be ineffective for fairness in $\mathcal{S}$. Therefore, to enhance fairness adaptation with the help of the auxiliary set $\mathcal{A}$, we propose to maximize the mutual information (MI) between the support set $\mathcal{S}$ and the auxiliary set $\mathcal{A}$. In consequence, the fairness adaptation on $\mathcal{S}$ will benefit from $\mathcal{A}$.
%%%%%However, due to the insufficiency of samples in $\mathcal{S}$, it remains non-trivial to leverage the auxiliary set for fairness adaptation on $\mathcal{S}$, since the auxiliary sets may not contain beneficial information for 
%First, it is difficult to utilize the auxiliary set. Due to the generalization gap between meta-training and meta-test resulting from the distribution difference, directly fine-tuning the model with the auxiliary set can lead to suboptimal performance. This is because the optimal fairness fine-tuning result cannot be guaranteed due to the existence of distribution differences. 
%Second, it is challenging to decide the weight of the auxiliary set in optimization. In particular, excessively focusing on the optimization with the auxiliary set can lead to reduced performance on the target meta-task. Nevertheless, assigning a lower weight can also result in less effectiveness of incorporating the auxiliary set. 

Generally, the support set $\mathcal{S}$ in $\mathcal{T}$ can be expressed as $\mathcal{S}=\{(x_1,y_1),(x_2,y_2),\dotsc,(x_{N\times K},y_{N\times K})\}$, which contains $K$ samples for each of $N$ classes. $x_i$ is an input sample, and $y_i$ is the corresponding label. We use $a_i\in\{0,1\}$ to denote its sensitive attribute. In particular, we propose to construct an auxiliary set that shares the same structure as the support set. In this way, the auxiliary set $\mathcal{A}$ can be represented as $\mathcal{A}=\{(x^*_1,y^*_1),(x_2^*,y^*_2),\dotsc,(x^*_{|\mathcal{A}|},y^*_{|\mathcal{A}|})\}$. Here $|\mathcal{A}|$, i.e., the size of the auxiliary set, is set as a controllable hyper-parameter. Moreover, based on the classification model $f(\cdot)$, we can obtain the sample embedding $\bx_i\in\mathbb{R}^d$, and the classification probabilities $\bp_i=f(x_i)\in\mathbb{R}^N$ for $x_i$. Here $d$ denotes the embedding dimension of samples, and $N$ is the number of classes in $\mathcal{T}$.
Particularly, we maximize the fairness-aware MI between $\mathcal{S}$ and $\mathcal{A}$ by
\begin{equation}
    \max_\theta I(\mathcal{S};\mathcal{A})=\max_{\theta}\sum\limits_{i=1}^{|\mathcal{S}|}\sum\limits_{j=1}^{|\mathcal{A}|} p(x_i,x^*_j;\theta)\log\frac{p(x_i|x_j^*;\theta)}{p(x_i;\theta)},
\end{equation}
where $\theta$ denotes the parameters of classification model $f(\cdot)$.
Since the MI term $I(\mathcal{S};\mathcal{A})$ is difficult to obtain and also intractable, it is infeasible to directly maximize it~\cite{oord2018representation}. Therefore, we first re-formulate the MI term to make it computationally tractable based on the property of conditional probabilities:
\begin{equation}
\begin{aligned}
I(\mathcal{S};\mathcal{A})&=\sum\limits_{i=1}^{|\mathcal{S}|}\sum\limits_{j=1}^{|\mathcal{A}|} p(x_i|x^*_j;\theta)p(x^*_j;\theta)\log\frac{p(x_i|x_j^*;\theta)}{p(x_i;\theta)}\\
    &=\sum\limits_{i=1}^{|\mathcal{S}|}\sum\limits_{j=1}^{|\mathcal{A}|} p(x_j^*|x_i;\theta)p(x_i;\theta)\log\frac{p(x_i|x_j^*;\theta)}{p(x_i;\theta)}.
    \label{eq:MI}
    \end{aligned}
\end{equation}
Since the support set $\mathcal{S}$ is randomly sampled, we can assume that the prior probability $p(x_i;\theta)$ follows a uniform distribution and set it as a constant: $p(x_i;\theta)=1/|\mathcal{S}|$, which thus can be ignored in optimization. Therefore, it remains to estimate $p(x_i|x^*_j;\theta)$ and $p(x_j^*|x_i;\theta)$ to obtain the value of $I(\mathcal{S};\mathcal{A})$.
\subsubsection{Estimation of $p(x_i|x^*_j;\theta)$} We first denote $\mathcal{S}_0$ and $\mathcal{S}_1$ as the sets of samples with sensitive attributes of $0$ and $1$, respectively\footnote{For the sake of simplicity, we focus on tasks with only binary sensitive attributes in this paper. Nevertheless, our work can be easily generalized to tasks with multiple types of sensitive attributes.}. In other words, $\mathcal{S}=\mathcal{S}_0\cup\mathcal{S}_1$ and $\mathcal{S}_0\cap\mathcal{S}_1=\emptyset$. Similarly, we define sets $\mathcal{A}_0$ and $\mathcal{A}_1$ for the auxiliary set $\mathcal{A}$. Then we propose to estimate $p(x_i|x^*_j;\theta)$ as follows:
\begin{equation}
    p(x_i|x^*_j;\theta)=\left\{\begin{aligned}\frac{\bp_i(y^*_j)}{\sum_{x_k\in \mathcal{S}_{a_i}}\bp_k(y^*_j)}&\ \ 
    \text{if}\ \ a_i=a_j^*,\\ 0\ \ \ \ \ \ \ \ \ \ \ &\ \ \text{else}.
    \end{aligned}\right.
\end{equation}
Here $\bp_i(y^*_j)\in\mathbb{R}$ denotes the classification probability of $x_i$ regarding $y^*_j$, which is the label of $x_j^*$.
Intuitively, the probability measures the alignment of the classification between the support sample $x_i$ and the auxiliary sample $x_j^*$, which (1) shares the same sensitive attribute with $x_i$ and (2) is also similar to $x_i$ regarding the classification output. In other words, maximizing $p(x_i|x^*_j;\theta)$ can increase the fairness adaptation consistency between sample $x_i$ and auxiliary samples that are specifically beneficial for the fairness adaptation with $x_i$, thus promoting the fairness adaptation performance.
\subsubsection{Estimation of $p(x_j^*|x_i;\theta)$} 
The term $p(x_j^*|x_i;\theta)$ in Eq. (\ref{eq:MI}) is conditioned on $x_i$ and denotes the probability of $x_j^*$ inferred by $x_i$. Moreover, since the value of $p(x_i|x^*_j;\theta)$ becomes zero when the sensitive attributes of $x_i$ and $x_j^*$ are different, we only need to estimate $p(x_j^*|x_i;\theta)$ when $x_i$ and $x_j^*$ share the same sensitive attributes, i.e., $a_i=a_j^*$. Therefore, since $x_i$ and $x_j^*$ maintain the same sensitive attributes, we can estimate the probability $p(x_j^*|x_i;\theta)$ based on the squared Euclidean distance between their embeddings without explicitly considering their fairness-aware correlation. In particular, we further normalize the probability with a softmax function to formulate term $p(x_j^*|x_i;\theta)$ as follows:

	    \begin{equation}
        \left.p(x_j^*|x_i;\theta)
        =\frac{\exp\left(-\|\bx_i- \bx_j^*\|_2^2\right)}{\sum_{x_k^*\in \mathcal{A}_{a_j^*}}\exp\left(-\|\bx_i- \bx_k^*\|_2^2\right)}
        \right.
        .
    \end{equation}
Furthermore, to ensure the consistency of sample representations in meta-training and meta-test data, we apply the $\ell_2$ normalization on both $\bx_i$ and $\bx_j^*$, which results in $\|\bx_i-\bx_j^*\|_2^2=2-2\bx_i^\top\cdot\bx_j^*$. In this manner, the logarithmic term $\log p(x_j^*|x_i;\theta)$ becomes:
\begin{equation}
\begin{aligned}
            \log\left(p(x_j^*|x_i;\theta)\right)
        &=\log\left(\frac{\exp\left(-2+2\bx_i^\top\cdot\bx_j^*\right)}{\sum_{x_k^*\in \mathcal{A}_{a_j^*}}\exp\left(-2+2\bx_i^\top\cdot\bx_k^*\right)}
        \right)\\
        &=2\bx_i^\top\cdot\bx_j^*-\log \sum_{x_k^*\in \mathcal{A}_{a_j^*}}\exp\left(2\bx_i^\top\cdot\bx_k^*\right).
\end{aligned}
\end{equation}
Finally, the MI loss $\mathcal{L}_{MI}$ can be derived as follows:
        \begin{equation}
        \begin{aligned}
\mathcal{L}_{MI}=&\frac{1}{|\mathcal{A}|} \sum\limits_{j=1}^{|\mathcal{A}|}\sum\limits_{x_i\in\mathcal{S}_{a_j^*}}
        -\frac{\bp_i(y^*_j)}{\sum_{x_k\in \mathcal{S}_{a_i}}\bp_k(y^*_j)}
        \left(2\bx_i^\top\cdot\bx_j^*\right.\\
        &\left.-\log\sum\nolimits_{x_k^*\in \mathcal{A}_{a_j^*}}\exp\left(2\bx_i^\top\cdot\bx_k^*\right)\right)
        .
        \end{aligned}
    \label{eq:mi_objective}
    \end{equation}
The overall fairness adaptation loss can be represented as the combination of fairness regularization terms on the support set $\mathcal{S}$ and the auxiliary set $\mathcal{A}$ along with the MI loss between $\mathcal{S}$ and $\mathcal{A}$:
\begin{equation}
    \mathcal{L}_{FA}=\mathcal{L}_{R}(\mathcal{S})+\gamma \left(\mathcal{L}_{R}(\mathcal{A}) +\mathcal{L}_{MI}\right),
    \label{eq:fa}
\end{equation}
	where $\gamma$ is an adjustable weight hyper-parameter to control the importance of the auxiliary set. Specifically, $\mathcal{L}_R$ denotes the regularized optimization loss:
 \begin{equation}
     \mathcal{L}_{R}(S)=\frac{1}{|\mathcal{S}|}\sum_{(x,y)\in\mathcal{S}}\ell(f(x),y)+\lambda R(\mathcal{S}),
 \end{equation}
where $\ell$ is the classification loss, and $R(\mathcal{S})$ denotes the fairness regularization term.
\vspace{-0.1in}
\subsection{Auxiliary Sets Selection}
The second problem of the generalization gap between meta-training and meta-test in fair few-shot learning can also pose a significant challenge in fairness adaptation. To address this issue, 
 %To tackle the second problem of imbalanced sensitive attributes in few-show fairness learning, 
 we propose to select the auxiliary set based on its similarity in fairness adaptation directions to the target meta-test task. 
In this way, incorporating the auxiliary set with a similar fairness adaptation direction can potentially leverage beneficial learned knowledge in meta-training to enhance fairness adaptation in the target meta-task.
 %In this way, the auxiliary set can improve the fairness adaptation performance that may be harmed by imbalanced sensitive attributes.
 %based on the fairness adaptation directions of support sets in meta-training tasks. This is because the imbalanced sensitive attributes can harm the performance of fairness adaptation, 
 %%%Intuitively, the auxiliary set should be beneficial for the fairness adaptation of the target meta-task. 
 However, it is difficult to identify the fairness adaptation direction of the auxiliary set that aligns with the target meta-task. It is possible that the auxiliary set holds a different or even opposite fairness adaptation direction from the target meta-task. As such, the incorporation of such an auxiliary set can even harm the fairness adaptation performance.
 Therefore, to select the auxiliary set with a similar fairness adaptation direction to the target meta-test task, we introduce a \textit{dynamic dictionary}, $\mathcal{A}_{can}$, which stores all candidate auxiliary sets for selection, with the keys being their corresponding fairness adaptation directions.    
 %%%propose to store all the candidate auxiliary sets in a \emph{dynamic dictionary} $\mathcal{A}_{can}=\{\mathcal{A}_1, \mathcal{A}_2, \dotsc, \mathcal{A}_{|\mathcal{A}_{can}|}\}$, where the keys are their fairness adaptation directions. 
 This allows us to efficiently identify and select an auxiliary set with a similar adaptation direction for the target meta-test task, thereby improving the fairness adaptation performance in the presence of the generalization gap.
 %%%In this way, given a target task along with its fairness adaptation direction, we can select an auxiliary set with similar adaptation directions from the dictionary. 

 Notably, this dictionary will be dynamically updated by adding a new auxiliary set after each meta-training step and meanwhile removing the oldest auxiliary set, of which the fairness adaptation direction is the most outdated.
 %%%older auxiliary sets that have less consistent fairness adaptation direction with the given new meta-test task. 
 In this manner, the dictionary also acts like a queue, which means that the size can be flexible and independent to fit various scenarios. Specifically, after each step on a meta-training task $\mathcal{T}=\{\mathcal{S},\mathcal{Q}\}$, we will enqueue the support set $\mathcal{S}$ as a candidate auxiliary set\footnote{Note that the auxiliary set size is controllable via randomly removing samples in $\mathcal{S}$ or incorporating new samples before enqueuing.} into $\mathcal{A}_{can}$ and remove the oldest auxiliary set. 
 The key of enqueued $\mathcal{S}$, which is the fairness adaptation direction of $\mathcal{S}$, is set as the gradient of $\mathcal{L}_R(\mathcal{S})$, i.e., $\nabla_\theta \mathcal{L}_R(\mathcal{S})$, where $\theta$ denotes the model parameters of $f(\cdot)$. 
 
\vspace{.05in}
\noindent\textbf{Identifying the true fairness adaptation direction.} With the help of the dynamic dictionary as a queue during meta-training, it may still remain difficult to obtain the fairness adaptation direction of the target meta-test task $\mathcal{T}$. This is because the fairness adaptation direction of $\mathcal{S}$ cannot faithfully reveal the true direction due to potentially imbalanced sensitive attributes. Therefore, to identify the true fairness adaptation direction without directly conducting fairness adaptation on the support set $\mathcal{S}$, we propose the use of a generator $g(\cdot)$, parameterized by $\phi$, to estimate the fairness adaptation results for each meta-test task. In particular, the generator $g(\cdot)$ takes the support set $\mathcal{S}$ as input and outputs an estimation of the gradient of $\mathcal{L}_R(\mathcal{S})$, i.e., $\nabla_\theta \mathcal{L}_R(\mathcal{S})$. To optimize the generator $g(\cdot)$, we introduce the Mean Squared Error (MSE) loss as the objective function as follows:
 \begin{equation}
       \mathcal{L}_E=\left\|g(\mathcal{S})- \nabla_\theta \mathcal{L}_R(\mathcal{S})\right\|_2^2,
       \label{eq:mse}
 \end{equation}
where $g(\mathcal{S})\in\mathbb{R}^{d_\theta}$ is the generator output, and $d_\theta$ is the size of the classification model parameter $\theta$.
 It is worth mentioning that the input of the generator $g(\cdot)$ is an entire support set $\mathcal{S}$, which means that the generator should be able to capture the contextual information within the support set. For this reason, we propose to leverage the transformer encoder architecture~\cite{vaswani2017attention} followed by a Multiple Layer Perceptron (MLP) as the implementation of the generator. In specific, the output of the generator can be expressed as:
 \begin{equation}
g(\mathcal{S})=\text{MLP}\left(\text{Mean}\left(\text{Transformer}\left(\mathbf{x}_1,\mathbf{x}_2,\dotsc,\mathbf{x}_{|\mathcal{S}|}\right)\right)\right).
\label{eq:generator}
 \end{equation}
 In this manner, the generator can estimate the corresponding fairness adaptation direction from $\mathcal{S}$, where the result can be used for selecting an auxiliary set.

After the meta-training process on a series of meta-training tasks $\{\mathcal{T}_1,\mathcal{T}_2, \dotsc, \mathcal{T}_T\}$, we can obtain a dictionary of candidate auxiliary sets in $\mathcal{A}_{can}=\{\mathcal{A}_1, \mathcal{A}_2, \dotsc, \mathcal{A}_{|\mathcal{A}_{can}|}\}$ along with their fairness adaptation directions as keys. Here we denote their corresponding keys as $\mathbf{k}(\mathcal{A})\in\mathbb{R}^{d_\theta}$. Then given a new meta-test task $\mathcal{T}_\text{test}=\{\mathcal{S}_\text{test}, \mathcal{Q}_\text{test}\}$, the corresponding selected auxiliary set $\mathcal{A}^*$ can be selected via the following criterion:
\begin{equation}
\mathcal{A}^*=\argmin_{\mathcal{A}\in\mathcal{A}_{can}}\text{dist}\left(g(\mathcal{S}_\text{test}), \mathbf{k}(\mathcal{A})\right),
\label{eq:select}
\end{equation}
where $\text{dist}(\cdot,\cdot)$ is a function to measure the distance between two vectors. In the experimentation, we implement it as the Euclidean distance. We can then efficiently select an auxiliary set from a significantly large dictionary based on the keys. It is noteworthy that to keep consistency between meta-training and meta-test, we will also select an auxiliary set for each meta-training task for optimization.

\begin{algorithm} [!t]
		\caption {Detailed training process of our framework.}
		\begin{algorithmic}[1]
			\REQUIRE Meta-training task distribution $P_{tr}$ from the meta-training data $\mathcal{D}_{tr}$, number of meta-training tasks $T$, number of fine-tuning steps $\tau$.
			\ENSURE A trained fairness-aware classification model $f(\cdot)$ and a generator model $g(\cdot)$.
			%// \text{Meta-training phase}
   \STATE Randomly initialize the dictionary queue $\mathcal{A}_{can}$;
			%\STATE Initialize the global model
                \FOR {$i=1,2,\dotsc,T$}
                \STATE Sample a meta-training task $\mathcal{T}_i=\{\mathcal{S},\mathcal{Q}\}\sim P_{tr}$;
                \STATE Obtain the fairness adaptation direction via Eq. (\ref{eq:generator});
                \STATE Select an auxiliary set $\mathcal{A}$ from the candidate auxiliary set dictionary $\mathcal{A}_{can}$ based on Eq. (\ref{eq:select});
                \FOR {$t=1,2,\dotsc,\tau$}
                   \STATE Conduct one step of fairness adaptation according to Eq.  (\ref{eq:fa}) and Eq. (\ref{eq:fine-tuning});
			\ENDFOR
                \STATE Meta-optimize classification model $f(\cdot)$ and generator $g(\cdot)$ based on Eq. (\ref{eq:meta1}) and Eq. (\ref{eq:meta2}), respectively;
                \STATE Enqueue support set $\mathcal{S}$ into the dictionary queue $\mathcal{A}_{can}$ and remove the oldest candidate auxiliary set in $\mathcal{A}_{can}$;
                \ENDFOR
		\end{algorithmic}
					\label{algorithm}
	\end{algorithm}

 \subsection{Meta-optimization}
 Our framework is optimized under the episodic meta-learning paradigm~\cite{finn2017model}. 
 Specifically, let $\theta$ denote the total parameters of the classification model $f(\cdot)$. In order to perform fairness adaptation, we first initialize the model parameters as $\theta_0\leftarrow\theta$. After that, given a specific meta-task $\mathcal{T}=\{\mathcal{S},\mathcal{Q}\}$, we conduct $\tau$ steps of gradient descent based on the fairness adaptation loss $\mathcal{L}_{FA}$ calculated on the support set $\mathcal{S}$.
Thus, the fairness adaptation process in $\mathcal{T}$ can be formulated as follows:
\begin{equation}
	    \theta_t \leftarrow \theta_{t-1} -\alpha\nabla_{\theta_{t-1}}\mathcal{L}_{FA}\left(\mathcal{S};\theta_{t-1}\right),
	    \label{eq:fine-tuning}
\end{equation}
	where $t\in\left\{1,2,\dotsc,\tau\right\}$ and $\mathcal{L}(\mathcal{S};\theta_{t-1})$ denotes the loss calculated based on the support set $\mathcal{S}$ with the parameters $\theta_{t-1}$. $\tau$ is the number of fine-tuning steps applied, and $\alpha$ is the learning rate in each fine-tuning step. After conducting $\tau$ steps of fine-tuning, we will meta-optimize the classification model $f(\cdot)$ with the loss calculated on the query set $\mathcal{Q}$. In specific, we meta-optimize the model parameters $\theta$ with the following update function:
 		\begin{equation}
	    \theta=:\theta-\beta_1\nabla_{\theta}\mathcal{L}_{FA}(\mathcal{Q};\theta_\tau),
     \label{eq:meta1}
     \end{equation}
 where $\beta_1$ is the meta-learning rate for the classification model $f(\cdot)$.

For the optimization of the generator $g(\cdot)$, parameterized by $\phi$, the update can be formulated as follows:
 \begin{equation}
     \phi=:\phi-\beta_2\nabla_{\phi}\mathcal{L}_E(\mathcal{S}; \theta_\tau),
          \label{eq:meta2}
 \end{equation}
 where $\mathcal{L}_E$ is the MSE loss introduced in Eq. (\ref{eq:mse}), and $\beta_2$ is the meta-learning rate for the generator $g(\cdot)$.
 In this way, the model parameters $\phi$ of $g(\cdot)$ will be updated based on loss $\mathcal{L}_E$ after the fairness adaptation of the classification model $f(\cdot)$. The detailed training process of our framework is demonstrated in Algorithm \ref{algorithm}.

%via the kNN algorithm.

\section{Experimental Evaluations}

\subsection{Datasets}
In this subsection, we introduce the datasets used in our experiments. To evaluate the performance of FEAST on fair few-shot learning, we conduct experiments on three prevalent real-world datasets: Adult~\cite{dua2017uci}, Crime~\cite{lichman2013uci}, and Bank~\cite{moro2014data}. The detailed dataset statistics are provided in Table~\ref{tab:statistics}.
\begin{itemize} [itemsep=5.0pt]
\item The Adult dataset contains information from 48,842 individuals from the 1994 US Census, where each instance is represented by 14 features and a binary label. Here the label indicates whether the income of a person is higher than 50K dollars. Following the data split setting in PDFM~\cite{zhao2020primal}, we split the dataset into 34 subsets based on the country information of instances. We consider gender as the sensitive attribute.
\item The Crime dataset includes information on 2,216 communities from different states in the U.S., where each instance consists of 98 features. Following~\cite{slack2020fairness}, the binary label of each instance is obtained by converting the continuous crime rate based on whether the crime rate of a community is in the top 50\% within the state. The sensitive attribute is whether African-Americans are among the highest or second highest populations in each community. We further split this dataset into 46 subsets by considering each state as a subset.
\item The Bank dataset consists of 41,188 individual instances in total. Specifically, each instance maintains 20 features along with a binary label that indicates whether the individual has subscribed to a term deposit. Here, we consider marital status as the binary sensitive attribute. Moreover, the dataset is split into 50 subsets based on the specific date records of instances.
\end{itemize} 
	\begin{table}
	\setlength\tabcolsep{8.0pt}%调列距

		\centering
		\renewcommand{\arraystretch}{1.3}

		\caption{Statistics of three real-world datasets. }
		\begin{tabular}{c|ccc}
		\hline
        \textbf{Dataset}&Adult&Crime&Bank\\
        \hline
        Sensitive Attribute &Gender&Race&Marital Status\\
        Label&Income&Crime Rate&Deposit\\
        \# Instances&48,482&2,216&41,188\\
        \# Features&12&98&17\\
        \# Subsets&34&46&50\\
        \# Training Subsets&22&30&40\\
        \# Validation Subsets&6&8&5\\
        \# Test Subsets&6&8&5\\
        \hline
		\end{tabular}
		\label{tab:statistics}

	\end{table}
\subsection{Experimental Settings}
To achieve a fair comparison of FEAST with competitive baselines, we conduct experiments with the state-of-the-art fair few-shot learning methods and other few-shot learning methods with fairness constraints. The details are provided below.
\begin{itemize}
    \item MAML~\cite{finn2017model}: This method utilizes a classic meta-learning framework to deal with the fair few-shot learning problem without explicitly applying fairness constraints.
    \item M-MAML~\cite{finn2017model}: This method uses the same framework as MAML while modifying datasets by removing the sensitive attribute of each instance to enhance fairness during optimization.
    \item Pretrain~\cite{zhao2020primal}: This method learns a single model on all meta-training data without episodic training. Moreover, a fairness constraint is added to the training objective. 
    \item F-MAML~\cite{zhao2020fair}: This method applies a fairness constraint in each episode and tunes a Lagrangian multiplier shared across different episodes for fair few-shot learning tasks.
    \item FM-dp and FM-eop (Fair-MAML)~\cite{slack2020fairness}: These two baselines provide a regularization term for each episode based on demographic parity (DP) and equal opportunity (EOP), respectively.
    \item PDFM~\cite{zhao2020primal}: This method leverages a primal-dual subgradient approach to ensure that the learned model can be fast adapted to a new episode in fair few-shot learning.
\end{itemize}
 Particularly, we use the average classification accuracy (ACC) over $T_\text{test}$ meta-test tasks to evaluate the prediction performance. For fairness performance, we propose to utilize demographic parity (DP) and equalized odds (EO), which are commonly used in existing works~\cite{chuang2021fair, zhao2020unfairness,dwork2012fairness,yurochkin2020training}. Since we consider the binary classification datasets, the output $f(x)\in\mathbb{R}$ denotes the prediction score of a specific sample $x$. In this manner, the metrics can be calculated over $T_\text{test}$ meta-test tasks sampled from the meta-test task distribution $P_{te}$ as follows:
\begin{equation}
    \Delta \text{DP}=\mathbb{E}_{\mathcal{T}\sim P_{te}}\left|\frac{1}{|\mathcal{Q}_0|}\sum\limits_{x\in \mathcal{Q}_0} f(x)-\frac{1}{|\mathcal{Q}_1|}\sum\limits_{x\in \mathcal{Q}_1} f(x)\right|,
\end{equation}
\begin{equation}
    \Delta \text{EO}=\mathbb{E}_{\mathcal{T}\sim P_{te}}\sum\limits_{y\in\{0,1\}}\left|\frac{1}{|\mathcal{Q}_0^y|}\sum\limits_{x\in \mathcal{Q}^y_0} f(x)-\frac{1}{|\mathcal{Q}_1^y|}\sum\limits_{x\in \mathcal{Q}^y_1} f(x)\right|,
\end{equation}
where $\mathcal{Q}_0$ and $\mathcal{Q}_1$ denote the query samples with a sensitive attribute of 0 and 1, respectively. Similarly, $\mathcal{Q}_0^y$ (or $\mathcal{Q}_1^y$) denotes the query samples in $\mathcal{Q}_0$ (or $\mathcal{Q}_1$) with label $y$. $P_{te}$ is the meta-test task distribution of meta-test sets $\mathcal{D}^n$. Our code is released at \href{https://github.com/SongW-SW/FEAST}{https://github.com/SongW-SW/FEAST}.
%%%%%We also provide experiments on the size of the auxiliary set $|\mathcal{A}|$ in Appendix.

	\begin{table*}[t]
		\setlength\tabcolsep{3.88pt}%调列距
	\footnotesize
		\centering
		\renewcommand{\arraystretch}{1.6}
		\caption{Results w.r.t. fairness and prediction performance of FEAST and baselines under different settings for all three datasets.}%%%%%%%%%%\kz{bold the best-performing method}
		\begin{tabularx}{\textwidth}{c|ccc|ccc|ccc|ccc|ccc|ccc}
			\hline
			%\multirow{2}{*}{Model}
			Dataset&\multicolumn{6}{c|}{\text{Adult}}&\multicolumn{6}{c|}{\text{Crime}}& \multicolumn{6}{c}{\text{Bank}}
			\\
			\hline
				Setting&\multicolumn{3}{c|}{5-shot}&\multicolumn{3}{c|}{10-shot}&\multicolumn{3}{c|}{5-shot}&\multicolumn{3}{c|}{10-shot}&\multicolumn{3}{c|}{5-shot}&\multicolumn{3}{c}{10-shot}\\\hline
    Metric&$\Delta$DP&$\Delta$EO&ACC&$\Delta$DP&$\Delta$EO&ACC&$\Delta$DP&$\Delta$EO&ACC&$\Delta$DP&$\Delta$EO&ACC&$\Delta$DP&$\Delta$EO&ACC&$\Delta$DP&$\Delta$EO&ACC\\\hline
MAML&0.473&0.706&0.801&0.409&0.584&\textbf{0.886}&0.558&0.952&0.718&0.443&0.832&0.792&0.214&0.573&\textbf{0.603}&0.185&0.496&0.619\\\hline
M-MAML&0.447&0.689&\textbf{0.826}&0.381&0.555&0.857&0.359&0.732&0.711&0.300&0.569&0.757&0.214&0.544&0.600&0.175&0.459&0.619\\\hline
F-MAML&0.339&0.432&0.825&0.310&0.353&0.840&0.503&0.871&0.719&0.463&0.707&0.762&0.207&0.585&0.575&0.181&0.528&\textbf{0.650}\\\hline
FM-dp&0.313&0.502&0.814&0.241&0.438&0.844&0.385&0.722&0.741&0.329&0.604&0.771&0.238&0.614&0.586&0.187&0.553&0.604\\\hline
FM-eop&0.430&0.703&0.812&0.370&0.601&0.846&0.352&0.706&0.739&0.311&0.591&0.804&0.289&0.683&0.581&0.245&0.600&0.640\\\hline
Pretrain&0.365&0.513&0.806&0.310&0.450&0.885&0.390&0.692&\textbf{0.746}&0.354&0.582&0.776&0.248&0.659&0.594&0.208&0.539&0.642\\\hline
PDFM&0.261&0.461&0.815&0.276&0.401&0.869&0.402&0.784&0.722&0.325&0.669&\textbf{0.816}&0.210&0.585&0.589&0.180&0.493&0.645\\\hline
FEAST&\textbf{0.258}&\textbf{0.355}&0.820&\textbf{0.235}&\textbf{0.256}&0.861&\textbf{0.203}&\textbf{0.309}&0.739&\textbf{0.164}&\textbf{0.217}&0.797&\textbf{0.190}&\textbf{0.524}&0.583&\textbf{0.154}&\textbf{0.414}&0.641\\\hline

		\end{tabularx}
		\label{tab:all_result}

	\end{table*}

%\newpage
		\begin{figure}[!t]
		\centering
		\subfigure{
\includegraphics[width=0.232\textwidth]{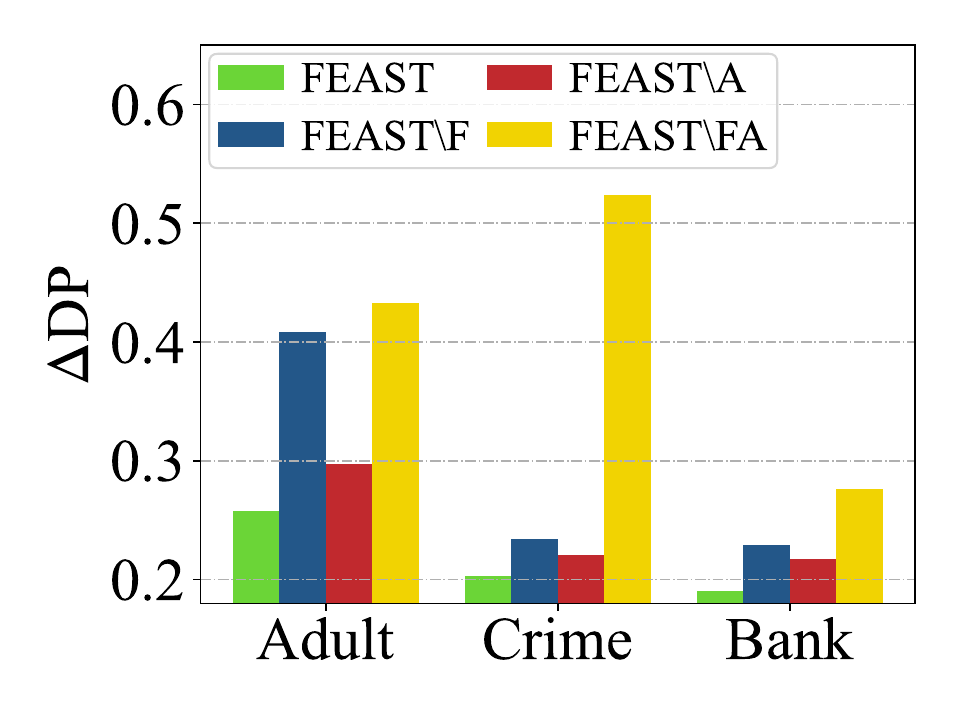}}
	\subfigure{
\includegraphics[width=0.232\textwidth]{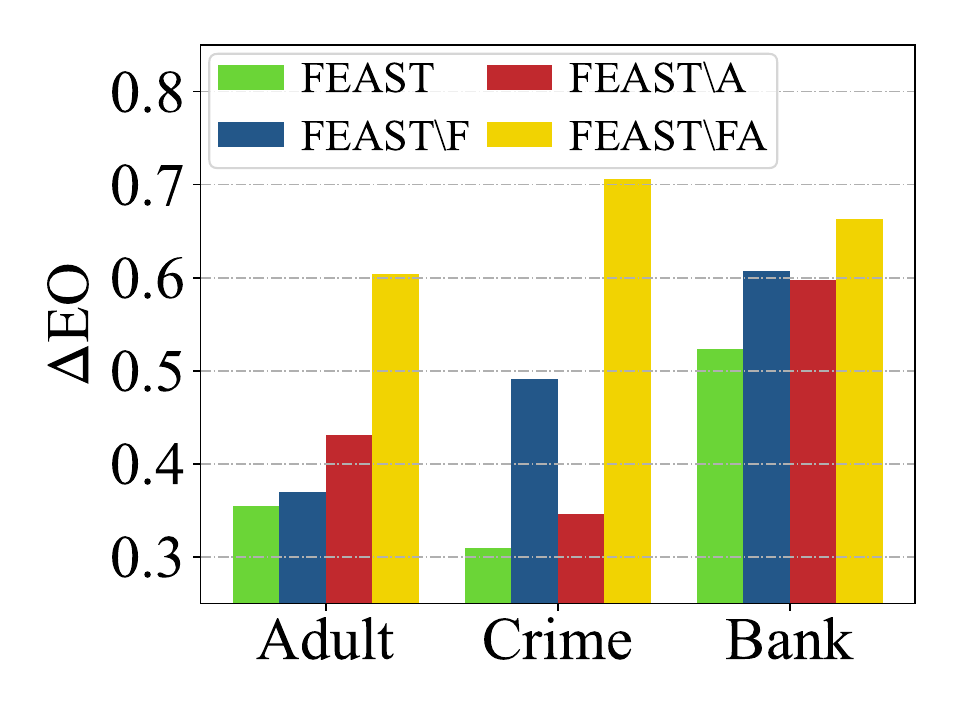}}
      \vspace{-0.12in}
    		\caption{Ablation study on our framework FEAST on three datasets under the 5-shot setting.}

		\label{fig:ablation}
	\end{figure}
\subsection{Performance Comparison}
Table~\ref{tab:all_result} presents the fairness and prediction performance comparison of FEAST and all other baselines on fair few-shot learning. Specifically, we report the results of $\Delta $DP, $\Delta $EO, and classification accuracy over 500 meta-test tasks for 10 repetitions. We conduct experiments on both 5-shot and 10-shot settings (i.e., $K=5$ and $K=10$). From Table~\ref{tab:all_result}, we can have following observations:
\begin{itemize}[itemsep=2.4pt]
    \item Our framework FEAST consistently outperforms other baselines in terms of fairness in all datasets under both 5-shot and 10-shot settings. These results provide compelling evidence for the effectiveness of our framework FEAST in fair few-shot learning.
    \item  The performance improvement of FEAST over other baselines is more significant on the Crime dataset. This is due to that in this dataset, each subset consists of fewer samples. Consequently, the learned fairness-aware meta-knowledge will be more difficult to be transferred in baselines. Nevertheless, our proposed fairness adaptation strategy based on mutual information can effectively deal with this scenario.
\item  The accuracy of FEAST is comparable with other baselines, demonstrating that FEAST can substantially reduce biases without sacrificing its classification capability. This is because our framework FEAST can select the auxiliary set with similar fairness adaptation directions and thus will not harm model performance regarding accuracy.
\item  FEAST is more robust to the changes of the number of support samples per class, i.e., when the number decreases from 10 to 5, FEAST has the least performance drop in comparison to other baselines. We believe this is primarily because, with fewer support samples, the problem of insufficient samples becomes more significant. Nevertheless, FEAST can effectively address this issue with the incorporation of auxiliary sets into fairness adaptation.
\end{itemize}

 		\begin{figure}[!t]
		\centering
		\subfigure{
\includegraphics[width=0.232\textwidth]{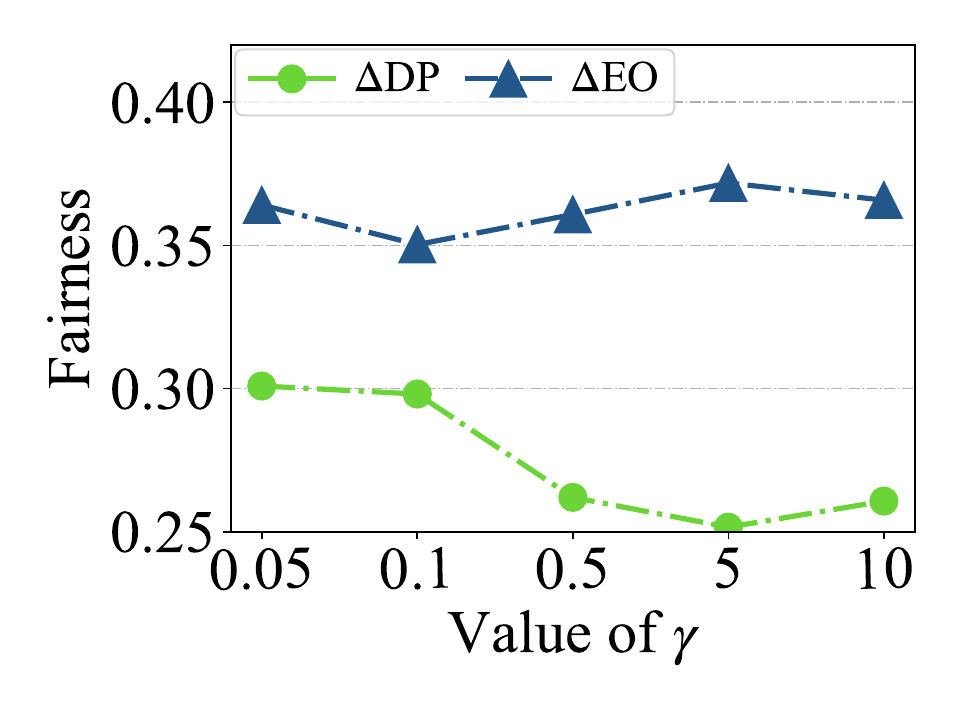}}
	\subfigure{
\includegraphics[width=0.232\textwidth]{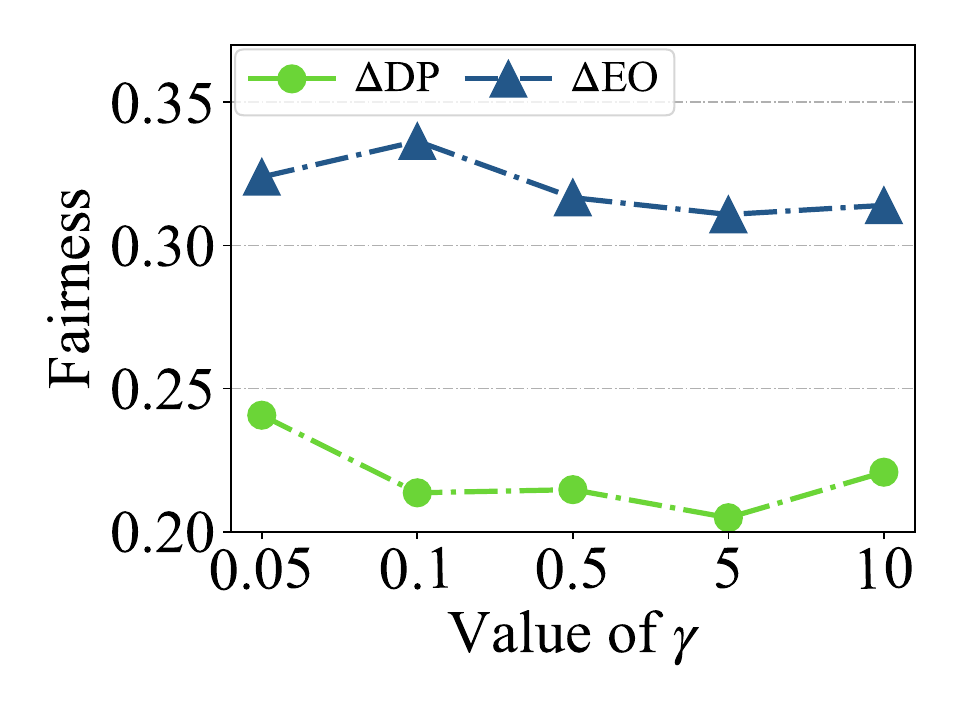}}
      \vspace{-0.12in}
		\caption{Results of FEAST on Adult (left) and Crime (right) with different values of $\gamma$.}
		\label{fig:gamma}
	\end{figure}
\subsection{Impact of Each Component in FEAST}
In this subsection, we conduct an ablation study on three datasets under the 5-shot setting to evaluate the effectiveness of different components in our framework by comparing FEAST with three degenerate versions: (1) FEAST without fairness adaptation based on MI, referred to as FEAST\textbackslash F. In this variant, the fairness adaptation process is simplified such that only fairness constraints are applied. (2) FEAST without auxiliary set selection, i.e., the auxiliary set is randomly sampled. We refer to this variant as FEAST\textbackslash A. (3) FEAST without both fairness adaptation and auxiliary set selection, referred to as FEAST\textbackslash FA. The results, as presented in Fig.~\ref{fig:ablation}, show that FEAST outperforms all other variants, validating the importance of both fairness adaptation and auxiliary set selection components in fair few-shot learning. Of particular interest is that the removal of the MI fairness adaptation has a more significant adverse impact on the Crime dataset, which contains significantly fewer meta-training samples. This result highlights the crucial role of this component in addressing the issue of insufficient training samples. 
In addition, when the two components are both removed, the fairness performance drops greatly. Such results indicate that the mutual impact brought by these two components is also critical for our proposed framework FEAST.
%%%%%without the auxiliary set selection, the performance of EO drops more significantly than that of DP. Such results indicate that the auxiliary set selection is more critical for fairness adaptation considering the labels of samples. 
%The reason is that Crime dataset consists of significantly fewer meta-training samples than the other two datasets, and thus maximally utilizing these samples for fairness adaptation becomes more crucial in this scenario. 
%%%%%Additionally, without the auxiliary set selection, the performance of FEAST drops more significantly in the 5-shot setting than in the 10-shot setting. Such results indicate that auxiliary set selection is critical for fairness adaptation, especially when there are fewer support samples per meta-task. 
%The result further indicates that the auxiliary set selection is critical for fairness adaptation, especially when each meta-task includes more support samples.

\subsection{Effect of Loss Weight $\gamma$}
 Given the significance of the auxiliary sets in the fairness adaptation, in this subsection, we further examine in-depth how the auxiliary sets will influence the performance of FEAST. Specifically, we vary the value of $\gamma$, which controls the importance of the auxiliary set loss during fairness adaptation. A higher value of $\gamma$ implies a larger importance weight on the auxiliary set and a smaller importance weight on the target task. Due to the limitation of space, we evaluate the model's performance on two datasets, Adult and Crime, using various values of $\gamma$ (similar results on the Bank dataset) on the 5-shot setting. The results, as shown in Fig.~\ref{fig:gamma}, indicate that a value around 0.5 for $\gamma$ generally yields better fairness performance for both datasets. This is mainly because a small $\gamma$ can be insufficient to leverage the fairness-aware meta-knowledge in auxiliary sets, while an excessively large value of $\gamma$ can result in the loss of crucial fairness information in the target meta-task. Moreover, the effect of different $\gamma$ values is more significant on the Adult dataset. The reason is that this dataset contains a larger number of samples in meta-training data. As a result, the learned fairness-aware knowledge is richer in the auxiliary sets, thus propagating the benefits from auxiliary sets.

\subsection{Effect of Auxiliary Set Size}
\begin{figure}[!t]
		\centering
		\subfigure{
\includegraphics[width=0.232\textwidth]{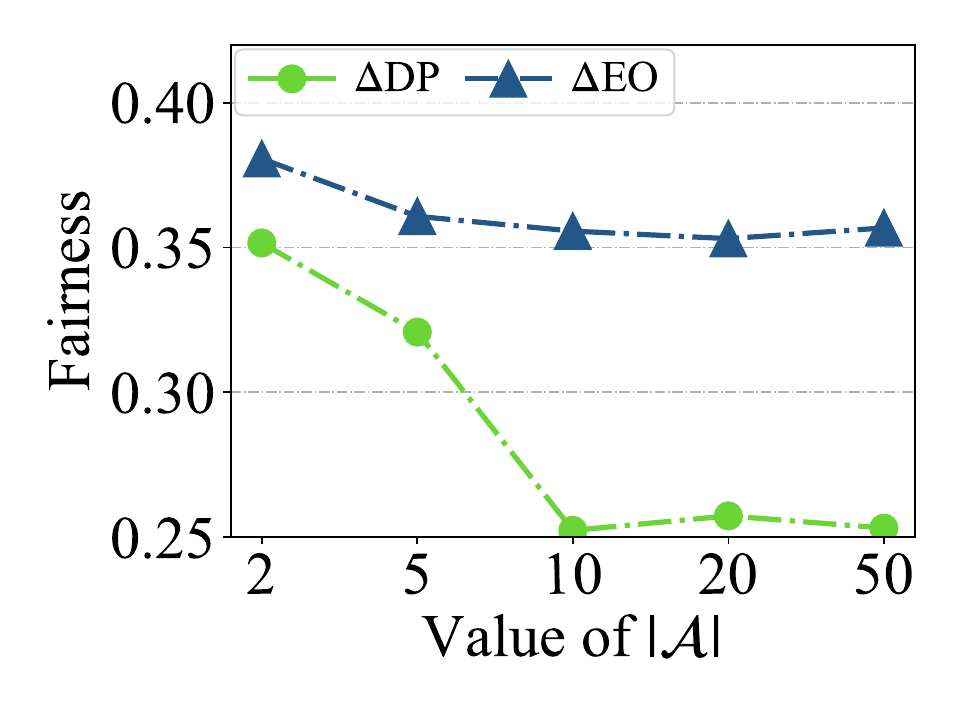}}
	\subfigure{
\includegraphics[width=0.232\textwidth]{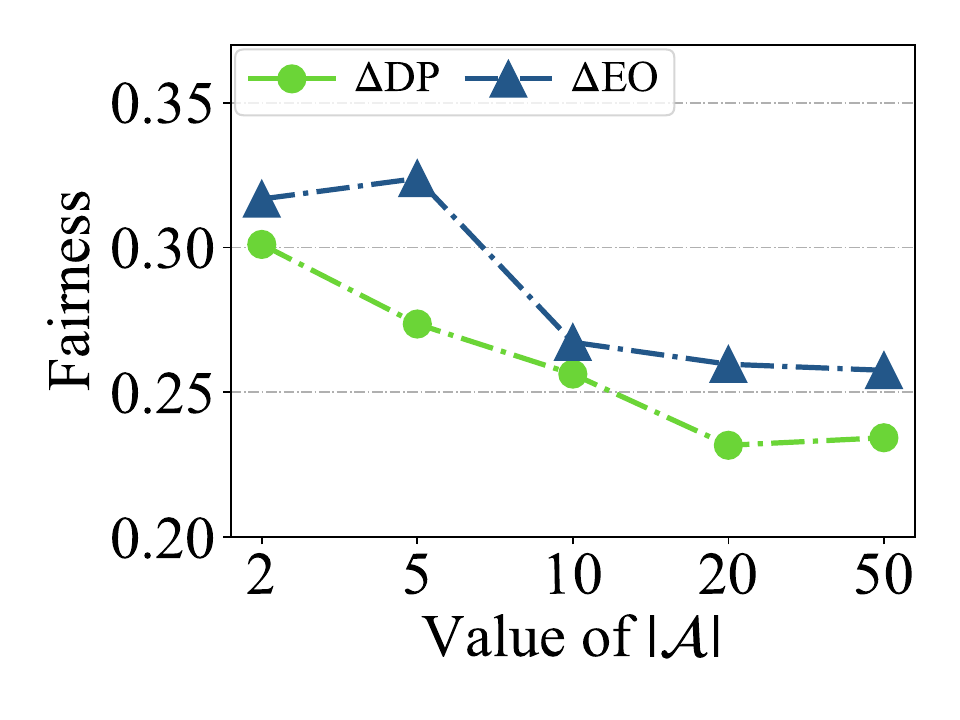}}
      \vspace{-0.12in}
		\caption{Results of FEAST on Adult under 5-shot (left) and 10-shot (right) settings with different values of $|\mathcal{A}|$.}
		\label{fig:aux}
	\end{figure}

In this section, we conduct experiments to evaluate the impacts brought by varying the size of the auxiliary set $\mathcal{A}$. Intuitively, the auxiliary set size $|\mathcal{A}|$ should be at least comparable with the support set, since an excessively small auxiliary set can be potentially insufficient for fairness adaptation. Specifically, we conduct experiments on dataset Adult under both 5-shot and 10-shot settings to evaluate the effect of auxiliary set size $|\mathcal{A}|$. From the results presented in Fig.~\ref{fig:aux}, we can make the following observations: (1) The fairness results are less satisfactory with a smaller value of $|\mathcal{A}|$, indicating that the capacity of $\mathcal{A}$ can be important in FEAST. With a small auxiliary set $\mathcal{A}$, the fairness adaptation effect will be reduced due to insufficient knowledge in $\mathcal{A}$. (2) When further increasing the size of $\mathcal{A}$, the fairness performance does not accordingly increase. This demonstrates that knowledge in a larger auxiliary set may not be helpful for fairness adaptation. (3) When the number of shots increases from 5 to 10, the best value of $|\mathcal{A}|$ also increases, implying that with a larger support set, the auxiliary set should also be expanded to provide more knowledge for fairness adaptation. In consequence, the fairness performance can be further improved.

\section{Related Work}
%In this section, we review related work from two aspects: few-shot learning and fairness-aware machine learning.

\subsection{Few-shot Learning}
    Few-shot learning aims to obtain satisfactory classification performance with only a few labeled samples as references~\cite{tian2020rethinking, tan2022transductive}. The typical approach is to accumulate transferable knowledge from meta-training tasks, which contain abundant labeled samples. Then such knowledge is generalized to meta-test tasks with limited labeled samples. Particularly, existing few-shot learning methods can be divided into two main categories: (1) \emph{Metric-based} methods propose to learn a metric function that matches samples in the query set with the support samples to conduct classification~\cite{liu2019learning,sung2018learning, wang2022faith,wang2022task}. For example, Prototypical Networks~\cite{snell2017prototypical} learn a prototype (i.e., the average embedding of samples in the same class) for each class and then classify query samples according to the Euclidean distances between query samples and each prototype. 
    Matching Networks~\cite{vinyals2016matching} output predictions for query samples via the similarity between query samples and each support sample. 
    (2) \emph{Optimization-based} methods aim to first fine-tune model parameters based on gradients calculated on support samples and then conduct meta-optimization on each meta-task~\cite{mishra2018simple,ravi2016optimization, wang2021reform, wang2022glitter}. As a classic example, MAML~\cite{finn2017model} learns a shared model parameter initialization for various meta-tasks with the proposed meta-optimization strategy. 
    LSTM-based meta-learner~\cite{ravi2016optimization} proposes an adjustable step size to update model parameters. 
\vspace{-0.05in}
\subsection{Fairness-aware Machine Learning}
Various fairness-aware algorithms have been proposed to mitigate the unwanted bias in machine learning models. Generally, there are two categories of statistical fairness notions: \emph{individual fairness} and \emph{group fairness}. In particular, individual fairness requires that the model results for similar individuals should also be similar~\cite{dwork2012fairness,yurochkin2020training,dong2023interpreting,dong2023fairness}. Here, the similarity between individuals can be measured via specific metrics (e.g., Euclidean distance) learned during training or from prior knowledge. On the other hand, group fairness refers to the statistical parity between subgroups (typically defined by sensitive attributes, e.g., gender and race) via specific algorithms~\cite{zemel2013learning,louizos2015variational,hardt2016equality, dong2022structural}. Common fairness learning tasks include fair classification~\cite{zafar2017fairness,feldman2015certifying}, regression~\cite{berk2017convex,calders2013controlling}, and recommendations~\cite{singh2018fairness}. Although these methods have demonstrated satisfactory performance in mitigating unfairness, it is noteworthy that existing works mainly focus on the settings where sufficient labeled samples are provided. As a result, it is challenging for these methods to accommodate few-shot scenarios with limited labeled samples. 

More recently, several methods are proposed to deal with the fair few-shot learning problem~\cite{slack2020fairness,zhao2020fair}. For example, PDFM~\cite{zhao2020primal} utilizes a primal-dual subgradient approach to ensure fast adaptation to a novel meta-task. In~\cite{zhao2020unfairness}, the authors propose to address fairness in supervised few-shot meta-learning models that are sensitive to discrimination in historical data by detecting and controlling the dependency effect of sensitive attributes on target prediction. Moreover, F-MAML~\cite{zhao2020fair} provides a fairness constraint for each episode and tunes a Lagrangian multiplier shared across different episodes based on a meta-learning mechanism.
However, these methods cannot effectively solve the problem of insufficient samples and the generalization gap. 
\vspace{-0.05in}
\section{Conclusion}
    In this paper, we propose a novel problem of fair few-shot learning, which focuses on 
    accurately and fairly predicting labels for samples in unseen data while using limited labeled samples as references. To tackle the challenges posed by insufficient samples and the generalization gap between meta-training and meta-test, we propose an innovative framework FEAST that utilizes learned fairness-aware meta-knowledge by incorporating auxiliary sets. In particular, our framework maximizes the mutual information between meta-tasks and the auxiliary sets to enhance fairness adaptation. Moreover, we select auxiliary sets based on the estimated fairness adaptation direction of meta-tasks to improve the fairness performance. %even when imbalanced sensitive attributes are present. 
    We conduct extensive experiments on three real-world datasets, and the results validate the superiority of FEAST over the state-of-the-art baselines. For future work, it is important to consider expanding the candidate auxiliary set with external knowledge, since samples in the dataset can be insufficient. In this case, incorporating external information for fairness adaptation can be crucial.

\newpage
\section{Acknowledgements}
	The work in this paper is supported by the National Science Foundation under grants (IIS-2006844, IIS-2144209, IIS-2223769, CNS2154962, and BCS-2228534), the Commonwealth Cyber Initiative
	awards (VV-1Q23-007 and HV-2Q23-003), the JP Morgan Chase
	Faculty Research Award, the Cisco Faculty Research Award, the Jefferson Lab subcontract 23-D0163, and the UVA 4-VA collaborative
research grant.

\bibliography{ecai}

\end{document}

\begin{frontmatter}
\title{Appendix}
\end{frontmatter}
\clearpage
\appendix
\section{Notations}
We provide the used notations in this paper along with their definitions for a more comprehensive understanding.
\begin{table}
%\small
\setlength\tabcolsep{8pt}
\caption{Notations used in this paper.} 
		\renewcommand{\arraystretch}{1.3}
\label{tb:symbols}
\begin{tabular}{cc}

\hline
\textbf{Notations}       & \textbf{Definitions or Descriptions} \\
\hline

$X$   &  the input feature\\
$Y$ & the input label\\
$A$ & the input sensitive attribute\\
$\mathcal{D}_{tr}$,$\mathcal{D}_{te}$ & the meta-training data and meta-test data\\
$\mathcal{T}$, $\mathcal{S}$, $\mathcal{Q}$&a meta-task and its support set and query set\\
% $\alpha_i$, $\beta_i$, $\tau_i$ & adaptation parameters for the $i$-th class\\
$N$&the number of support classes in each meta-task\\
$K$&the number of labeled samples in each class\\
$\mathcal{A}$&the auxiliary set\\
$\mathcal{A}_{can}$&the candidate auxiliary sets\\
% $N_t$, $K_t$& the value of $N$ and $K$ during meta-training\\
%%$\mathbf{s}_i$&the embedding of the $i$-th class in each meta-task\\
%%$\mathbf{q}_i$&the embedding of the $i$-th query node in each meta-task\\
%%$\mathbf{p}_i$&the classification probabilities of the $i$-th query node over 
$T$&the number of meta-training tasks\\
$T_{test}$&the number of meta-test tasks\\
$\tau$& the number of fairness adaptation steps\\
$f(\cdot)$&the classification model\\
$g(\cdot)$&the generator model\\
$\alpha$&the learning rate for fairness adaptation\\
$\beta_1$&the meta-learning rate for $f(\cdot)$\\
$\beta_2$&the meta-learning rate for $g(\cdot)$\\
\hline
\end{tabular}

\end{table}
\section{Reproducibility}
\subsection{Implementation Details}
In this section, we provide details on the implementation settings of our framework FEAST. 
	\label{appendix:implementation}
	Specifically, we implement FEAST with PyTorch~\cite{paszke2017automatic} and train our framework with a single 48GB Nvidia A6000 GPU. For the encoder model, we leverage a three-layer MLP, where the hidden sizes are set as 40, 40, and 2, respectively. The activation function of ReLU is utilized. During training, the learning rate of $\beta_1$ and $\beta_2$ are both set as 0.001, while $\alpha$ is set as 0.01. For optimization, we leverage the Adam~\cite{kingma2014adam} optimization strategy with the weight decay rate set as $10^{-4}$. During the meta-test process, we randomly sample 500 meta-test tasks from the meta-test data $\mathcal{D}_{te}$ with a query set size $|\mathcal{Q}|$ of 10. It is noteworthy that to satisfy the requirements of computing the fair metric, we ensure that there are both sensitive attributes (i.e., $a=0$ and $a=1$) in the query set. In order to preserve consistency for fair comparisons, we keep the meta-test tasks identical for all baselines. The number of meta-training tasks $T$ is set as 500. The number of fairness adaptation steps is set as 10. The loss weight $\gamma$ is set as 0.5. For the size of the auxiliary set $\mathcal{A}$, we set it as 10 and 20 for 5-shot and 10-shot settings, respectively. 

\subsection{Required Packages}

The more detailed package requirements are listed as below.
	\begin{itemize}
	    \item Python == 3.7.10
	    \item torch == 1.8.1
	    \item numpy == 1.18.5
        \item scipy == 1.5.3
        \item cuda == 11.0
    \item tensorboard == 2.2.2
    \item scikit-learn == 0.24.1
    \item pandas==1.2.3
	\end{itemize}
\subsection{Baseline Settings}
	Here, we present the detailed parameter settings of the baselines used in our experiments. We mainly follow the original setting in the corresponding source code while adopting specific selections of parameters for better performance. For baselines that are not originally proposed for fair few-shot learning, we use the same parameter settings as our framework for consistency.
\begin{itemize}
    \item MAML~\cite{finn2017model}: For MAML, we set the meta-learning rate as 0.001 and the learning rate as 0.01.
    \item M-MAML~\cite{finn2017model}: For this baseline, we use the same parameter setting as MAML.
    \item Pretrain~\cite{zhao2020primal}: For this baseline, we set the learning rate as 0.005.
    \item F-MAML~\cite{zhao2020fair}: For this baseline, we follow the setting in the source code and set the learning rate as 0.001 with an inner learning rate of 0.01.
    \item FM-dp and FM-eop (Fair-MAML)~\cite{slack2020fairness}: For these two baselines, we follow the settings in the source code and set the meta-learning rate as 0.001.
    \item PDFM~\cite{zhao2020primal}: We follow the setting in the source code and set the meta-learning rate as 0.001 and the meta-dual learning rate as 0.01. The inner learning rate is set as 0.01. The fairness bound is set as 0.05.
\end{itemize}
\bibliography{ecai}